\documentclass[letterpaper, 10 pt, conference]{ieeeconf}  

\IEEEoverridecommandlockouts                              

\usepackage{graphics} 
\usepackage{epsfig} 
\usepackage{times} 
\usepackage{amsmath} 
\usepackage{amssymb}  
\usepackage{makecell}
\usepackage{bbding}
\usepackage{pifont}
\usepackage{hyperref}

\usepackage{ulem}       
\usepackage{subfig}     
\usepackage{booktabs}   
\usepackage{multirow}   
\usepackage{tabularx}   
\usepackage[table]{xcolor}     

\newlength\savewidth

\title{\LARGE \bf
MovSAM: A Single-image Moving Object Segmentation Framework Based on Deep Thinking}

\author{Chang Nie\textsuperscript{1}, Yiqing Xu\textsuperscript{2}, Guangming Wang\textsuperscript{3}, Zhe Liu\textsuperscript{1}, Yanzi Miao\textsuperscript{2}, and Hesheng Wang\textsuperscript{1} 
\thanks{This work was supported in part by the Natural Science Foundation of China under Grant 62225309, U24A20278, 62361166632 and U21A20480. (Corresponding Author: Hesheng Wang)
}
\thanks{\textsuperscript{1}Department of Automation,
Key Laboratory of System Control and Information Processing of Ministry of
Education, Key Laboratory of Marine Intelligent Equipment and System of
Ministry of Education, Shanghai Engineering Research Center of Intelligent
Control and Management, Shanghai Jiao Tong University, Shanghai 200240,
China. }
\thanks{\textsuperscript{2}The Advanced Robotics Research Center, Artificial Intelligence Research Institute and School of Information and Control Engineering, China University of Mining and Technology, Xuzhou, 221116, China.}
\thanks{\textsuperscript{3}Department of Engineering, University of Cambridge, Cambridge CB2 1PZ, U.K. }
}

\begin{document}
\maketitle
\thispagestyle{empty}
\pagestyle{empty}
	
\begin{abstract}
Moving object segmentation plays a vital role in understanding dynamic visual environments. While existing methods rely on multi-frame image sequences to identify moving objects, single-image MOS is critical for applications like motion intention prediction and handling camera frame drops. However, segmenting moving objects from a single image remains challenging for existing methods due to the absence of temporal cues. To address this gap, we propose MovSAM, the first framework for single-image moving object segmentation. MovSAM leverages a Multimodal Large Language Model (MLLM) enhanced with Chain-of-Thought (CoT) prompting to search the moving object and generate text prompts based on deep thinking for segmentation. These prompts are cross-fused with visual features from the Segment Anything Model (SAM) and a Vision-Language Model (VLM), enabling logic-driven moving object segmentation. The segmentation results then undergo a deep thinking refinement loop, allowing MovSAM to iteratively improve its understanding of the scene context and inter-object relationships with logical reasoning. This innovative approach enables MovSAM to segment moving objects in single images by considering scene understanding. We implement MovSAM in the real world to validate its practical application and effectiveness for autonomous driving scenarios where the multi-frame methods fail. Furthermore, despite the inherent advantage of multi-frame methods in utilizing temporal information, MovSAM achieves state-of-the-art performance across public MOS benchmarks, reaching 92.5\% on J\&F. Our implementation will be available at \url{https://github.com/IRMVLab/MovSAM}.

\end{abstract}

\section{Introduction}
Moving Object Segmentation (MOS) is a fundamental task for autonomous driving and robot perception, focusing on the pixel-level segmentation of the moving object in images. Current MOS methods predominantly rely on analyzing multiple frames or optical flow to segment the moving object \cite{xie2024moving, cho2023treating}. These methods use temporal cues, comparing successive images to detect motion.

However, when autonomous vehicles or robots are gaming the environment, the motion intentions of other objects are difficult for these multi-frame methods to infer from subtle movement changes. Furthermore, the motion of the autonomous system itself can lead to misinterpret the movements of other objects. Multi-frame techniques also struggle with occlusions, where objects are partially hidden, as pixel changes become unreliable. In contrast, humans can effectively identify potential moving objects in a single image through logical scene understanding. This capability highlights the demand for single-image MOS solutions, which could enhance safety in camera-failure scenarios. Beyond safety, accurately discerning motion from single images is also vital for emerging applications such as realistic video generation, where understanding object moving trends is paramount for image-based video synthesis.

\begin{figure}[t]
  \centering
   \includegraphics[width=0.85\linewidth]{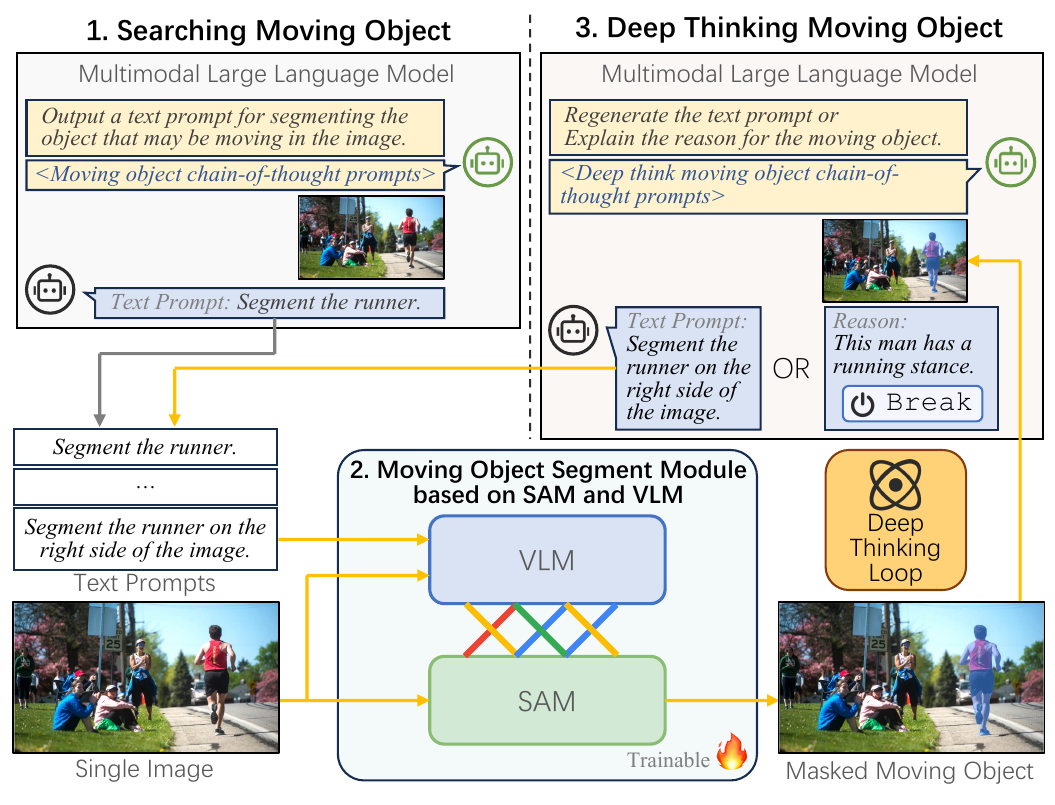}
    \vspace{-0.2cm}
   \caption{\textbf{Overview of MovSAM deep thinking about segmenting the moving object from a single image.} MovSAM begins with the MLLM searching for the moving object to generate an descriptive text prompt. Subsequently, the text prompt and the image enter a thinking loop: the moving object is first segmented by SAM and VLM in the segmentation module. Next, MLLM is deep thinking the segmentation result. The text prompt is regenerated if the segmentation is incorrect. If the segmentation is correct, the loop breaks, outputting the final segmented moving object.}
   \vspace{-0.4cm}
   \label{fig:overall}
\end{figure}

For logical reasoning, recent large models show significant progress. Multimodal Large Language Model (MLLM), evolving from Large Language Model (LLM), now exhibits the capacity to reason about visual content. Despite these advancements, directly applying MLLM to the single-image moving object segmentation (IMOS) task remains challenging due to their limited inherent segmentation capabilities.

For the image segmentation problem, the Segment Anything Model (SAM) \cite{kirillov2023segment} demonstrates great zero-shot segmentation ability. However, SAM primarily focuses on image texture analysis and lacks deep semantic scene comprehension. To enrich SAM with semantic awareness, recent efforts have integrated it with Vision-Language Model (VLM). VLM excels at understanding scenes and establishing logical connections between text and images. Combining VLM with SAM enables a more comprehensive scene interpretation. Nevertheless, simply combining these models can still lead to inaccuracies arising from model hallucinations or flawed reasoning. These problems cause contradictions between predicted and expected segmentation outcomes, thereby hindering precise moving object segmentation.

To overcome these limitations, we introduce MovSAM, a novel framework designed for deep thinking to segment moving objects from a single image. MovSAM leverages the strengths of MLLM, SAM, and VLM in a synergistic manner. As illustrated in Fig. \ref{fig:overall}, MovSAM initially employs an MLLM with chain-of-thought (CoT) prompting for searching. This allows for step-by-step analysis of the image to identify moving objects and generate a descriptive text prompt to guide segmentation. The CoT is a key technique for enabling deep thinking in advanced models, like GPT-o1 and Deepseek-R1 \cite{guo2025deepseek}. Subsequently, MovSAM integrates this text prompt and the input image, fusing SAM and a tailored VLM for the segmentation process. This fusion allows SAM to benefit from the relational scene understanding provided by the VLM, facilitating moving object segmentation based on logical inference from a single image. To mitigate potential reasoning errors, MovSAM incorporates a deep thinking loop. This loop iteratively evaluates the plausibility of segmented object using CoT prompts and provides a natural language rationale. If the segmentation is deemed incorrect, MovSAM refines the segmentation prompts and the segmentation results through iterative thinking, which mimics human-like reasoning processes.

Our main contributions are as follows:
\begin{itemize}
    \item To the best of our knowledge, MovSAM for the first time builds a deep thinking pipeline to realize single-image moving object segmentation without reliance on temporal information, which can bridge the gap of multi-frame methods.
    
    \item MovSAM introduces a deep thinking loop to effectively guide step-by-step logical reasoning on single images, addressing challenges like occlusion and motion blur. Furthermore, MovSAM incorporates a feature aggregation network to improve the fusion of SAM and VLM features for enhanced moving object segmentation.

    \item Real-world driving scenes implementations verifies the application and effectiveness of MovSAM for scenarios where the multi-frame methods fail. Moreover, MovSAM demonstrates state-of-the-art performance on MOS datasets, reaching 92.5\% J\&F on DAVIS2016. Notably, despite using only single-image, MovSAM outperforms existing multi-frame methods that utilize additional temporal information.

\end{itemize}

\begin{figure*}[t]
  \centering
   \includegraphics[width=0.85\linewidth]{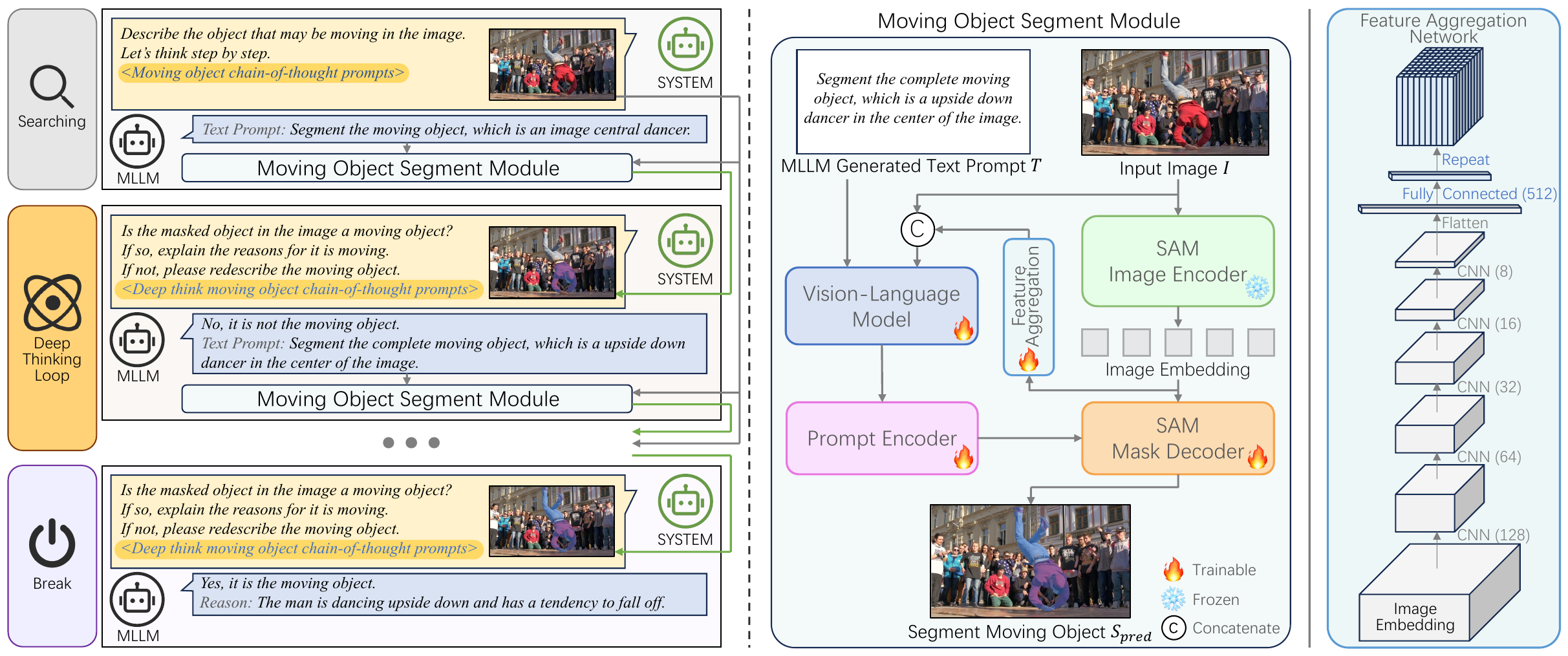}
    \vspace{-0.2cm}
   \caption{\textbf{The pipeline of the proposed MovSAM.} First, Multimodal Large Language Model (MLLM) generates a text prompt with the moving object CoT, describing the moving object in the image (Sec. \ref{init}). Subsequently, in the moving object segment module, MovSAM cross-fuses the features of SAM and VLM to logically reason about the image to segment the moving object (Sec. \ref{segment}). The segmentation is then thought by the MLLM in a deep thinking loop (Sec. \ref{evaluate}). If the segmentation is incorrect, a new text prompt is generated for segmentation again. If correct, MLLM explains the reasons for the movement. Through this closed-loop framework of guidance, segmentation, and deep thinking, MovSAM achieves single-image moving object segmentation based on scene understanding and logical reasoning.}
   \vspace{-0.6cm}
   \label{fig:pipeline}
\end{figure*}

\vspace{-0.1cm}
\section{Related Work}
\vspace{-0.1cm}
\textbf{Moving Object Segmentation (MOS)} addresses the problem of identifying and segmenting dynamic objects at the pixel level. This capability is crucial for applications such as 3D reconstruction, robots gaming with the environments, and collaborative robotics. Existing MOS methods predominantly rely on motion cues derived from optical flow or multi-frame video sequences. Optical flow-based techniques leverage short-term motion information to distinguish moving objects from static background \cite{cho2023treating}. However, generating accurate optical flow typically necessitates multiple frames. Video-based methods, on the other hand, establish relationships between visual appearance and motion patterns \cite{voigtlaender2019feelvos, koh2017primary}. Despite their advancements, these existing methods generally require either optical flow computations or temporal information from multiple frames. Consequently, they struggle to infer motion from a single image, limiting their applicability in scenarios where frame drops occur or gaming with the environment.

\textbf{Segment Anything Model (SAM)} has emerged as a foundational model for image segmentation \cite{wang2024sam}. Trained on extensive datasets and employing a transformer architecture, SAM has shown promise in diverse fields, including medical image segmentation and object tracking. However, SAM primarily operates with point or bounding box prompts for segmentation. Directly incorporating text prompts remains a challenge for SAM. To address this, LISA \cite{lai2024lisa} integrates MLLM with SAM for instructed segmentation. While these methods enhance SAM with text, they tend to focus on recognizing object categories and lack the deep thinking ability required for more complex segmentation tasks. Thus, tasks demanding intricate logical inference, such as IMOS, still present considerable challenges.

\textbf{Vision Language Model (VLM)} is designed to process both visual and textual information, enabling them to tackle a wide range of downstream tasks. Pre-training VLM on large-scale datasets significantly improves their ability to generalize to new situations. For instance, BEiT-3 \cite{wang2022image} stands as a multimodal pre-training foundation model applicable to various tasks, including visual reasoning, visual question answering, and image captioning.

\textbf{Multimodal Large Language Model (MLLM)} represents a further evolution from Large Language Model (LLM), extending into the multimodal domain. Compared to general VLM, MLLM often exhibits a stronger emphasis on textual processing and is more complex to adapt due to their substantial number of parameters. Models like Llama 3.2 Vision excels in tasks such as image captioning and visual grounding, accurately identifying and describing objects in images based on natural language instructions.

These breakthroughs suggest a pathway towards segmenting moving objects directly from single images. Despite these promising developments, significant challenges still exist in tasks requiring complex reasoning about motion from static visual information. Our work bridges this gap by proposing a deep thinking framework, achieving complex reasoning on an image for moving object segmentation.

\vspace{-0.1cm}
\section{Methodology}
\vspace{-0.2cm}
\subsection{System Framework}
\vspace{-0.2cm}
MovSAM addresses the task of segmenting moving objects in images by combining three key components: an MLLM, SAM, and a VLM. As illustrated in Fig. \ref{fig:pipeline}, MovSAM first uses the MLLM with the CoT prompts to search the moving object. The MLLM is asked to generate a text prompt $T$ describing the moving object in the image, following a series of complex reasoning steps.

Then, this text prompt $T$ is used with the image $I$ for moving object segmentation. First, the image features are encoded into an image embedding by the image encoder of SAM. Secondly, the image embedding is extracted into global image features by a feature aggregation network and concatenated to input image. VLM analyzes the high-dimensional image features based on the text prompt and outputs multimodal features to the prompt encoder for encoding. Finally, the SAM mask decoder generates the segmentation mask, identifying the moving object in the image.

To refine the segmentation, MovSAM incorporates a deep thinking loop driven by the MLLM. If the MLLM determines the segmentation to be unsatisfactory, a new prompt is generated, and the segmentation process is repeated. This iterative refinement continues until the MLLM judges the segmentation to be correct, at which point the loop breaks. The MLLM then interprets the successfully segmented moving object and outputs the final segmentation result $S_{pred}$.

\vspace{-0.1cm}
\subsection{Searching Moving Object via MLLM with CoT}
\label{init}
\vspace{-0.2cm}
MovSAM utilizes the reasoning capabilities of an MLLM to search the moving object to generate an initial text prompt. MLLM, having been trained on vast amounts of visual and textual data, is capable of complex logical inference about images. In MovSAM, we employ a capable MLLM, such as Llama 3.2 Vision \cite{dubey2024llama} (although other advanced MLLMs could also be used). We leverage the pre-existing knowledge of the MLLM to search the moving object through complex reasoning. In this way, the MovSAM is completely free of human involvement and automatically describes the moving object. Thus, at the searching of the MovSAM process, the MLLM generates a text prompt $T$ designed to facilitate the segmentation of the moving object within the input image $I$. This prompt generation is guided by a Chain-of-Thought (CoT) approach.

Chain-of-Thought (CoT) is a technique that encourages language models to solve problems by following a series of logical steps. This step-by-step approach enhances the model's ability to arrive at effective solutions. TABLE \ref{tab:move_cot} illustrates an example of the CoT prompts used in MovSAM to guide the MLLM in describing moving objects. These prompts systematically direct the MLLM to analyze the image scene. Initially, the MLLM is prompted to describe the relationships between objects within the image. The CoT then provides general characteristics typically associated with moving objects. Finally, the MLLM is instructed to identify and describe the moving object, culminating in the generation of the summarized text prompt $T$. This text prompt $T$ is subsequently passed to the segmentation module for single-image moving object segmentation.

\vspace{-0.1cm}
\subsection{Moving Object Segment Module}
\vspace{-0.2cm}
\label{segment}
Given the text prompt $T$ generated by the MLLM and the image $I$, the moving object segmentation module in MovSAM predicts a segmentation mask $S_{pred}$ that delineates the moving object. This process can be represented as:
\vspace{-0.4cm}
\begin{equation}
S_{pred}  = \mathit{\Phi}(T, I),
\vspace{-0.4cm}
\end{equation}
where $\mathit{\Phi}$ represents the segmentation network within MovSAM, which integrates SAM and a VLM. In contrast to the MLLM, which is primarily used for prompt generation and reasoning, the network $\mathit{\Phi}$ is designed to be adaptable and suitable for fine-tuning.

\begin{table}[t]\footnotesize 
\centering
\caption{\textbf{Example of moving object CoT prompts.}}
\vspace{-6pt}
\setlength{\tabcolsep}{1.0mm}
\renewcommand\arraystretch{1.1}
\begin{tabular}{c}
\toprule
\begin{tabular}[c]{@{}l@{}}
\textbf{Step 1}: Observe the objects in the image and their relationships.\\ 
\textbf{Step 2}: Pay attention to any possible moving object.\\ 
\textbf{Step 3}: Generally, moving objects exhibit certain characteristics, such as:\\ 
\textbf{(a)} Non-rigid objects may have a flowing effect, like hair or clothing.\\ 
\textbf{(b)} Edges of objects may show motion blur.\\ 
\textbf{(c)} An object may be suspended in the air without support.\\ 
\textbf{(d)} If an object has minimal support (like a toe touching the ground),\\it typically indicates that motion is about to occur.\\ 
\textbf{Step 4}: Identify based on, but not limited to, these features.\\ 
\textbf{Step 5}: If there is one, provide a text prompt $T$ for segmenting the\\moving object, including its category, position, and the motion it is\\undertaking. For example: “A red ball is flying in the air, in the upper\\right corner of the frame.”\\ 
\textbf{Step 6}: If there is no moving object, output: “No moving object.”\end{tabular}\\ \bottomrule
\end{tabular}
\label{tab:move_cot}
\vspace{-0.7cm}
\end{table}

To effectively extract image features relevant for moving object segmentation, MovSAM employs SAM2 \cite{ravi2024sam} as its core segmentation network. SAM excels at analyzing image textures; however, it inherently lacks a comprehensive understanding of image semantics.

To imbue SAM with semantic awareness, MovSAM integrates BEiT-3 \cite{wang2022image} alongside SAM2. This integration allows for the simultaneous processing of both the text prompt $T$ and the image $I$. Specifically, MovSAM retains the original SAM image encoder and mask decoder, as these components are well-suited for our single-image moving object segmentation framework.

Recognizing that the image embedding extracted by the SAM pre-trained image encoder contains rich image features, we leverage these features to enhance the image analysis capabilities of the VLM. MovSAM incorporates a feature aggregation network to consolidate global image features. As depicted in Fig. \ref{fig:pipeline}, this network consists of five CNN layers and a fully connected layer, designed to map the image embedding to a 512-dimensional global feature vector. This global feature vector is then concatenated with each pixel of the image embedding, enriching the image representation.

Subsequently, this enhanced image representation, along with the text prompt $T$, is input to the VLM to extract multimodal features. These high-dimensional multimodal features from the VLM are then fed into a modified prompt encoder. Finally, the encoded prompt features and the image embedding are processed by the mask decoder to produce the final segmentation of the image. Through this architecture, MovSAM achieves moving object segmentation by systematically reasoning about the image and effectively fusing the feature extraction strengths of SAM and the semantic understanding of the VLM.

\vspace{-0.1cm}
\subsection{Deep Thinking Moving Object}
\label{evaluate}
\vspace{-0.2cm}
To mitigate potential inaccuracies or illusions, MovSAM incorporates a deep thinking loop. This loop serves to evaluate and justify the segmentation. Specifically, MovSAM re-inputs the segmented image region back into the MLLM. The deep thinking moving object CoT is used to guide this evaluation, as shown in TABLE \ref{tab:eva_cot}.

\begin{table}[t]\footnotesize 
\centering
\caption{\textbf{Example of deep thinking moving object CoT prompts.}}
\vspace{-6pt}
\setlength{\tabcolsep}{1.0mm}
\renewcommand\arraystretch{1.1}
\begin{tabular}{c}
\toprule
\begin{tabular}[c]{@{}l@{}}
\textbf{Step 1}: Analyze the relationships between objects in the image\\to  understand their spatial arrangement.\\ 
\textbf{Step 2}: Identify the object covered by the blue mask, focusing\\on the region it encompasses.\\ 
\textbf{Step 3}: Determine if the masked object is a moving object by\\assessing its characteristics.\\ 
\textbf{Step 4}: Evaluate whether the masked object is completely\\segmented, checking for any missed areas.\\ 
\textbf{Step 5}: Assess if the masked object includes any extraneous\\objects that are not moving.\end{tabular}\\ \bottomrule
\end{tabular}
\label{tab:eva_cot}
\vspace{-0.7cm}
\end{table}

MovSAM instructs the MLLM to rigorously assess whether the segmented object genuinely exhibits characteristics of movement. If the MLLM identifies the segmentation as incorrect, it proceeds to refine the text prompt, aiming for greater accuracy in subsequent moving object segmentations. This iterative process forms a deep thinking loop. Within this loop, the MLLM continuously refines the text prompt based on its evaluation. Thus, the segmentation module can benefit from these improved prompts, enhancing segmentation accuracy through logical reasoning.

Conversely, if the MLLM confirms the segmentation as accurate, MovSAM then prompts the MLLM to articulate the reasoning behind classifying the object as moving. Upon providing a satisfactory explanation, the deep thinking loop breaks. This framework allows MovSAM to fully exploit the reasoning capabilities of the MLLM, leading to high-quality, supervised segmentation outcomes.

\begin{figure*}[t]
  \centering
   \includegraphics[width=0.78\linewidth]{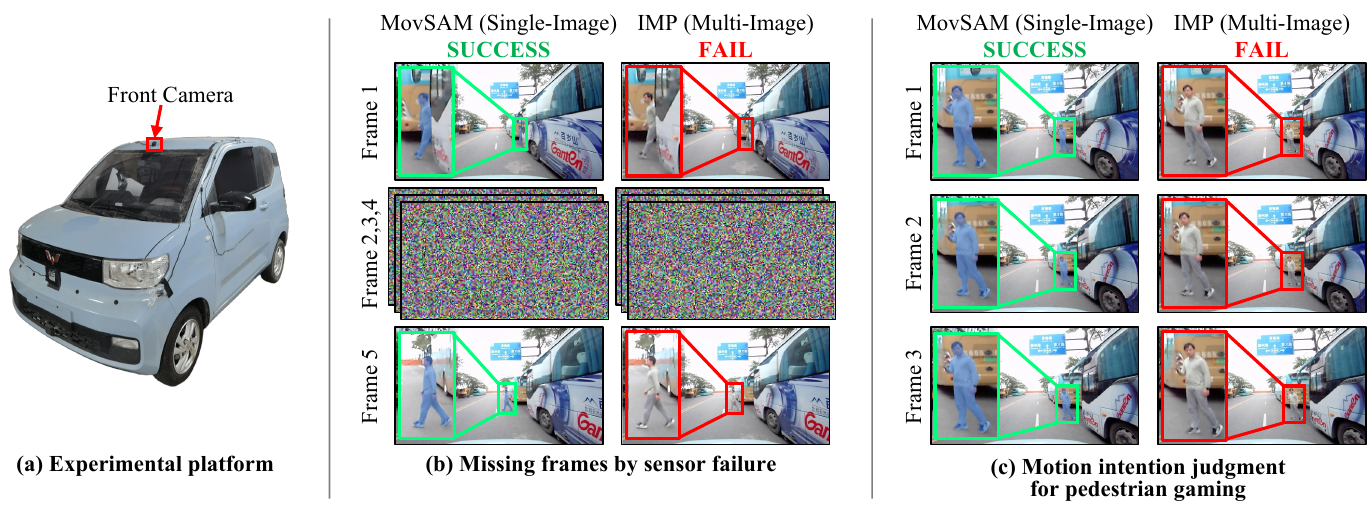}
    \vspace{-0.4cm}
   \caption{\textbf{Real-world driving scenes.} Sensor failures leading to dropped frames and pedestrian gaming scenes are very difficult for multi-image methods. In contrast, MovSAM based on deep thinking can accurately segment the moving object.}
   \vspace{-0.6cm}
   \label{fig:real}
\end{figure*}

\vspace{-0.1cm}
\subsection{Training and Loss Function}
\vspace{-0.2cm}
During the training phase, we designate the VLM and the feature aggregation network as trainable components. This design choice enables the network to effectively learn and extract relevant features from both image and text modalities. Consequently, the VLM gains sufficient information to accurately locate moving objects within images. For SAM, we adopt a strategy of training all modules except for the image encoder. We maintain the original SAM image encoder due to its pre-established high capacity for feature extraction, derived from extensive pre-training. Prior to commencing training, we initialize MovSAM with publicly available pre-trained weights from SAM-ViT-Huge \cite{kirillov2023segment} and BEIT-3-Large \cite{wang2022image}. The MLLM employed in our framework is Llama-3.2-11B-Vision \cite{dubey2024llama}.

To train MovSAM, we utilize a combination of two commonly used loss functions: Dice Loss and Binary Cross-Entropy (BCE) Loss, each assigned an equal weight. Dice Loss provides a measure of global segmentation performance. It is calculated as:
\vspace{-0.4cm}
\begin{align}
\mathcal{L}_{Dice} = 1 - \frac{2 \sum_{i=1}^{N} p_i y_i}{\sum_{i=1}^{N} p_i + \sum_{i=1}^{N} y_i},
\vspace{-2.4cm}
\end{align}
where $N$ is the total number of pixels in the image $I$, $y_i$ is the ground truth label of pixel $i$-th sample (either 0 or 1), and $p_i$ is the predicted value for the $i$-th pixel. 

where $N$ represents the total number of pixels in the image $I$. $y_i$ is the ground truth label for the $i$-th pixel (either 0 or 1), and $p_i$ is the predicted probability value for the $i$-th pixel.

BCE Loss operates at the pixel level, assessing the accuracy of each pixel's classification. It is defined as:
\vspace{-0.2cm}
\begin{align}
\mathcal{L}_{BCE} = -\frac{1}{N} \sum_{i=1}^{N} \left[ y_i \log(p_i) + (1 - y_i) \log(1 - p_i) \right].
\vspace{-1cm}
\end{align}

By integrating both Dice Loss and BCE Loss, we enhance the overall robustness of MovSAM during training. The combined loss function for MovSAM is thus:
\vspace{-0.2cm}
\begin{align}
\mathcal{L}_{MovSAM} = \mathcal{L}_{BCE}(S_{pred}, S_{gt}) + \mathcal{L}_{Dice}(S_{pred}, S_{gt}).
\vspace{-1cm}
\end{align}
Here, the \( S_{pred} \) is the predicted segmentation mask generated by MovSAM. The \( S_{gt} \) is the corresponding ground truth segmentation mask.

For training data, we collect a dataset by manually filtering images from the DAVIS2016 \cite{perazzi2016benchmark}, FBMS \cite{ochs2013segmentation}, and SegTrackV2 \cite{li2013video} datasets. Our filtering criterion ensured that only images in which the moving object is clearly distinguishable are included in the training set. The MovSAM framework is trained for 100 epochs, utilizing four RTX 8000 GPUs. The framework is based on PyTorch. MovSAM can perform up to five deep thinking loops.

\vspace{-0.1cm}
\section{Real-World Applications}
\vspace{-0.2cm}

We collect data on the real-world experimental platform shown in Fig. \ref{fig:real} (a) to simulate challenging real-world conditions, including sensor failure and pedestrian gaming. When the camera malfunctions or communication blockages cause dropped frames, the system may only have single images available. Fig. \ref{fig:real} (b) demonstrates that multi-image based IMP \cite{lee2022iteratively} fails to segment the moving object. However, MovSAM based on deep thinking can successfully segment the moving object with only one image. 

Furthermore, when the autonomous vehicle is gaming with objects in the environment, accurately understanding the motion intention of the objects is very important to ensure safety. Fig. \ref{fig:real} (c) shows a scenario where a pedestrian is stationary but might start moving. Such a common scene cannot be addressed by multi-image based IMP. Instead, MovSAM can accurately determine the motion trend of the pedestrian by reasoning about their current state. These real-world experiments demonstrate the practical application of MovSAM and its ability to address limitations of earlier methods. This unique advantage stems from the framework, which enables deep reasoning about images.

\vspace{-0.1cm}
\section{Large-Scale Validation Experiments}
\vspace{-0.2cm}
\subsection{Implementation Details}
\vspace{-0.2cm}
The performance of MovSAM is rigorously evaluated through comparisons with multi-frame Video Object Segmentation (VOS) methods on large-scale datasets. Notably, MovSAM uses only single-frame images, which means that the information obtained by MovSAM is less than other multi-frame methods.

In alignment with standard evaluation practices \cite{cho2023treating, wang2021swiftnet}, region accuracy is measured using the Jaccard index ($\mathcal{J}$), which is a common metric in image segmentation tasks. The $\mathcal{J}$ represents the overlap between ground truth $S_{gt}$ and predicted segmentation masks $S_{pred}$:
\vspace{-0.2cm}
\begin{equation}
\mathcal{J} = \frac{\left| S_{gt} \cap S_{pred} \right|}{\left| S_{gt} \cup S_{pred} \right|}.
\vspace{-0.2cm}
\end{equation}

Contour accuracy is assessed using the average boundary F-measure ($\mathcal{F}$):
\vspace{-0.2cm}
\begin{equation}
\mathcal{F} = \frac{2 \times P \times R}{P + R}, P = \frac{\left| S_{gt} \cap S_{pred} \right|}{\left| S_{pred} \right|}, R = \frac{\left| S_{gt} \cap S_{pred} \right|}{\left| S_{gt} \right|},
\vspace{-0.1cm}
\end{equation}
where Precision ($P$) is correctly predicted foreground / total predicted foreground. Recall ($R$) is correctly predicted foreground / ground truth foreground. $\mathcal{J\&F}$ is the average of $\mathcal{J}$ and $\mathcal{F}$, a segmentation performance measure.

\vspace{-0.2cm}
\subsection{Quantitative Results}
\vspace{-0.1cm}
\textbf{DAVIS2016:} TABLE~\ref{tab:Quantitative Results} presents quantitative results on DAVIS2016 dataset.
MovSAM attains state-of-the-art performance while using only a single-frame image. MovSAM is distinctive as it does not require optical flow estimation or multi-frame inputs. This capability is attributed to the iterative refinement process of MovSAM, which facilitates the step-by-step identification of moving objects within a single image.

\begin{table}[ht]\footnotesize 
\centering
\vspace{-0.2cm}
\caption{\textbf{Quantitative results} on DAVIS2016 and FBMS datasets. OF indicates Optical Flow, MI means Multi-Image.}
\vspace{-0.2cm}
\setlength{\tabcolsep}{1.3mm}
\renewcommand\arraystretch{1.0}
\begin{tabular}{l|cc|ccc|c}
\toprule
\multirow{2}{*}{Method}  & \multirow{2}{*}{OF} & \multirow{2}{*}{MI} & \multicolumn{3}{c|}{DAVIS2016} & \multicolumn{1}{c}{FBMS} \\ \cline{4-7} 
                         & & & $\mathcal{J\&F}$ ↑       & $\mathcal{J}$ ↑        & $\mathcal{F}$ ↑        & $\mathcal{J}$ ↑   \\ \hline\hline 
D$^2$Conv3D \cite{schmidt2022d2conv3d}       & \ding{55}                          & \ding{51}             &  86.0     &  85.5  & 86.5   & -   \\
IMP \cite{lee2022iteratively}           &  \ding{55}                         & \ding{51}             &  85.6     &  84.5  & \underline{86.7}   & 77.5    \\
TMO \cite{cho2023treating}             & \ding{51}              & \ding{55}             &  86.1    &  85.6  & 86.6   & 79.9    \\ 
FlowP-SAM+FlowI-SAM \cite{xie2024moving}        & \ding{51}                & \ding{51}             &  \underline{86.7}     &  \underline{87.7}  & 85.6   & \underline{82.8}    \\ \hline
\textbf{MovSAM}        & \ding{55}                          & \ding{55}   & \textbf{92.5}      & \textbf{90.4}   &  \textbf{94.6}  & \textbf{83.9}  \\ \bottomrule
\end{tabular}
\label{tab:Quantitative Results}
\vspace{-0.3cm}
\end{table}

\textbf{FBMS:} The quantitative evaluation on FBMS dataset reveals that MovSAM achieves state-of-the-art performance in TABLE~\ref{tab:Quantitative Results}. While TMO \cite{cho2023treating} utilizes optical flow for segmentation, MovSAM achieves superior results on the FBMS dataset. This success is facilitated by the effective feature fusion between SAM and VLM, enabled by its feature aggregation network.

\textbf{YouTube-Objects:} TABLE \ref{tab:YouTube-Objects} summarizes the evaluation across different categories within the YouTube-Objects dataset. MovSAM achieves state-of-the-art (SOTA) performance in most categories without training on this dataset. This highlights the effectiveness of the iterative refinement process of MovSAM in improving segmentation results and enabling accurate segmentation of complex moving objects.

\begin{table}[ht]\footnotesize
\centering
\vspace{-0.2cm}
\caption{\textbf{Quantitative results} on the YouTube-Objects dataset. The $\mathcal{J}$ ↑ is reported.}
\vspace{-0.2cm}
\setlength{\tabcolsep}{0.55mm}
\renewcommand\arraystretch{1.0}
\begin{tabular}{l|cccccccc|c}
\toprule
\multirow{2}{*}{Method} & \multicolumn{8}{c|}{Categories} & \multirow{2}{*}{Mean}  \\ \cline{2-9} 
                     & Aeroplane & Bird & Boat & Car & Cat & Cow & Dog & Horse &  \\ \hline\hline 
                        
WCS-Net \cite{zhang2020unsupervised} & 81.8 & \underline{81.1} & 67.7 & 79.2 & 64.7 & 65.8 & 73.4 & \underline{68.6} & 72.8 \\
RTNet \cite{ren2021reciprocal} & 84.1 & 80.2 & \underline{70.1} & 79.5 & 71.8 & 70.1 & 71.3 & 65.1 & 74.0 \\
AMC-Net \cite{yang2021learning} & 78.9 & 80.9 & 67.4 & \textbf{82.0} & 69.0 & 69.6 & 75.8 & 63.0 & 73.3 \\
TMO \cite{cho2023treating}& \underline{85.7} & 80.0 & \underline{70.1} & 78.0 & \underline{73.6} & \underline{70.3} & \underline{76.8} & 66.2 & \underline{75.1} \\
\hline
\textbf{MovSAM} & \textbf{89.4} & \textbf{82.8}  & \textbf{76.8} & \underline{80.6} & \textbf{81.2} & \textbf{74.7} & \textbf{77.7} & \textbf{69.0} & \textbf{79.0} \\
\bottomrule
\end{tabular}
\label{tab:YouTube-Objects}
\vspace{-0.4cm}
\end{table}

\vspace{-0.2cm}
\subsection{Qualitative Results}
\begin{figure}[t]
  \centering
   \includegraphics[width=0.99\linewidth]{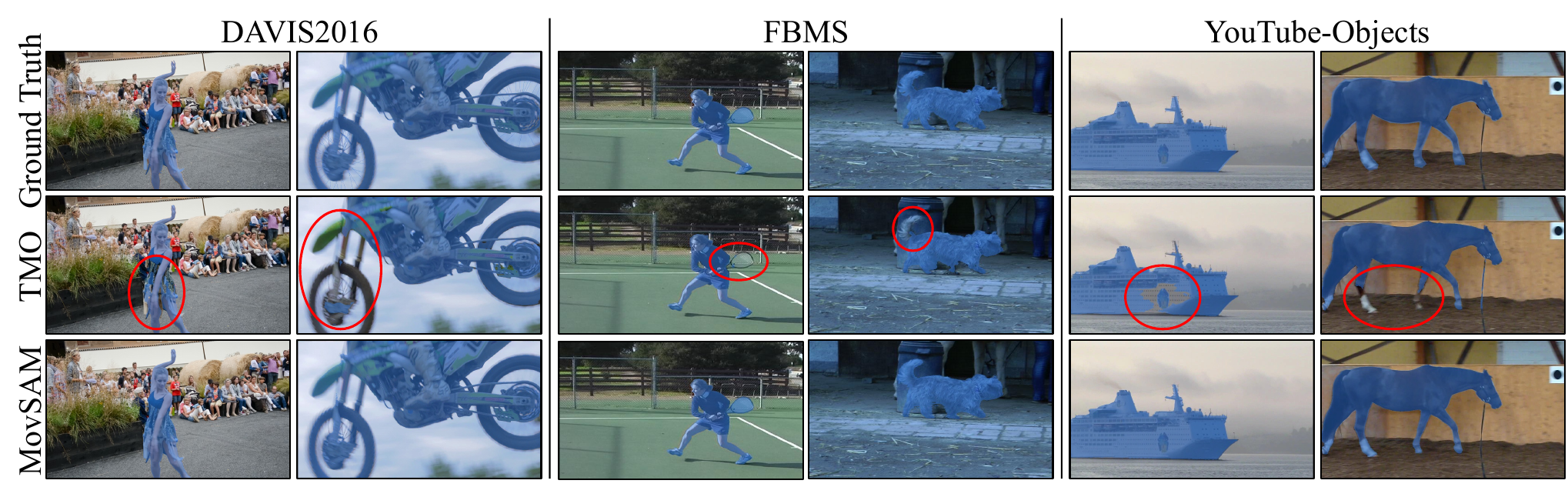}
    \vspace{-0.7cm}
   \caption{\textbf{The qualitative results of TMO and MovSAM on various MOS datasets.} The major errors are indicated by red circles. Motion blur can be addressed by deep thinking.}
   \vspace{-0.4cm}
   \label{fig:Qualitative_result}
\end{figure}

\vspace{-0.2cm}
\textbf{MOS datasets:} Fig. \ref{fig:Qualitative_result} shows the qualitative comparison between MovSAM and TMO \cite{cho2023treating}, conducted on the DAVIS2016, FBMS and YouTube-Objects datasets. Despite using optical flow information, TMO still faces segmentation errors in some detailed segments, such as the dance-twirl sequence in the DAVIS2016 dataset and the tennis sequence in the FBMS dataset. In contrast, MovSAM is better at capturing these details. This improvement is due to the strong image segmentation ability of SAM in MovSAM, combined with the use of motion cues within segmentation.

\begin{figure}[t]
  \centering
   \includegraphics[width=0.90\linewidth]{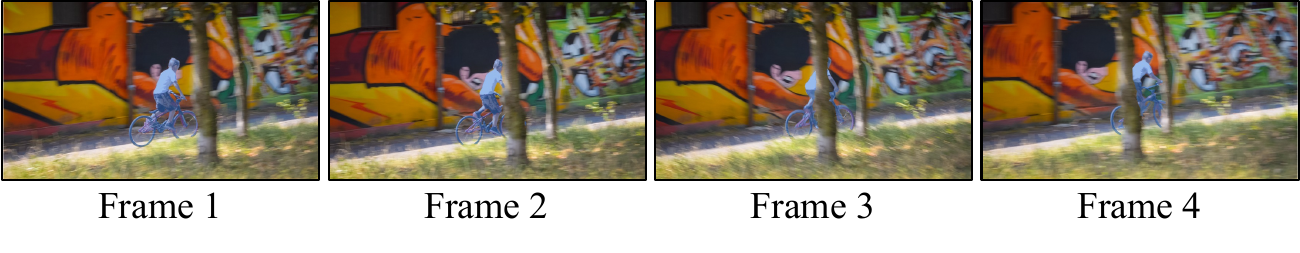}
    \vspace{-0.4cm}
   \caption{\textbf{The sequence results of MovSAM on bmx-trees in the occluded scene.} MovSAM can still segment the object under occlusion for each image of the sequence.}
   \vspace{-0.4cm}
   \label{fig:seq}
\end{figure}

\textbf{Sequence with occlusion:} As shown in Fig. \ref{fig:seq}, we further examine the capability of MovSAM in handling visual obstructions. MovSAM can segment the moving object very well for each frame without multi-frame information. Even in partially occluded scenes, MovSAM can still segment the object completely. Notably, this task was previously solved by combining multiple frames or optical flow. MovSAM, however, achieves robust segmentation using only single-frame information, showing its superior capability for processing moving objects with occlusion.

\begin{figure}[t]
  \centering
   \includegraphics[width=0.90\linewidth]{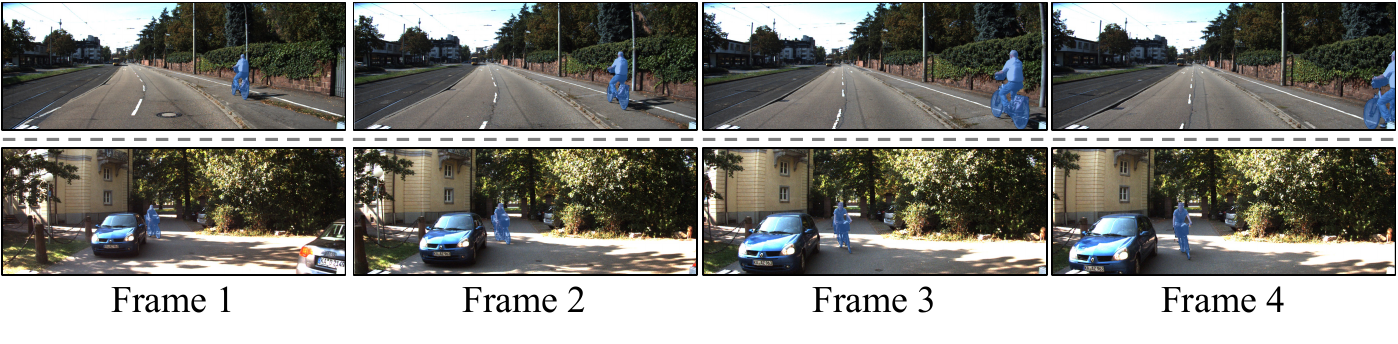}
    \vspace{-0.3cm}
   \caption{\textbf{The sequence results of MovSAM on the KITTI dataset.} This shows the application of MovSAM in autonomous driving scenes.}
   \vspace{-0.6cm}
   \label{fig:kitti}
\end{figure}

\textbf{KITTI:} We conducted experiments directly on the KITTI dataset with each single image for autonomous driving scenes, as shown in Fig. \ref{fig:kitti}. In such a completely unlearned scenario, MovSAM can still accurately segment the moving object. This illustrates the strong cross-domain capability of MovSAM. This adaptability is attributed to the deep thinking approach in MovSAM to analyze the relationships between objects within a scene.

\vspace{-0.2cm}
\subsection{Ablation Study}
\vspace{-0.2cm}
We analyzed the effectiveness of the MovSAM modules through ablation studies. The experimental setup and evaluation metrics are consistent with those described above.

\textbf{The effects of the feature aggregation network} are tested in TABLE~\ref{tab:Ablation study} (a). The performance of MovSAM decreases without the feature aggregation network. The feature aggregation network efficiently encodes high-dimensional SAM image embeddings, thus providing rich image features for VLM. This also proves that MovSAM effectively achieves the feature cross-fusion between SAM and VLM.

\textbf{The effects of the deep thinking loop} are tested in TABLE~\ref{tab:Ablation study} (b).
The deep thinking loop is designed to check the correctness and completeness of the segmentation results. The absence of the deep thinking loop prevents MLLM and the segmentation module from optimizing the prompts and segmentation results together. This leads to lower performance of MovSAM when facing complex scenes.

\begin{figure}[t]
  \centering
   \includegraphics[width=0.80\linewidth]{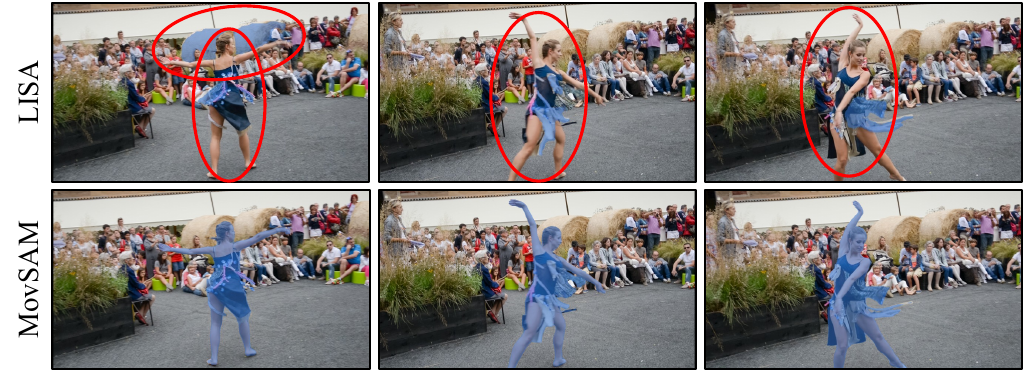}
    \vspace{-0.2cm}
   \caption{\textbf{Comparisons with language-guided segmentation method.} The state-of-the-art language-guided segmentation method, LISA, cannot correctly segment the dancer. This suggests that single-image moving object segmentation needs specialized frameworks to be addressed, rather than being tackled by generic methods.}
   \vspace{-0.4cm}
   \label{fig:LISA}
\end{figure}

\textbf{Comparisons with language-guided segmentation method} are tested in TABLE~\ref{tab:Ablation study} (c). The SOTA language-guided segmentation method, LISA \cite{lai2024lisa} without fine-tuning, performs poorly because it fails to learn the characteristics of moving objects, such as fuzzy edges and motion illusions, shown in Fig. \ref{fig:LISA}. Moreover, LISA with fine-tuning lacks deep thinking about the segmented object, causing either misidentify the moving object or fail to segment it completely. In contrast, MovSAM, with its deep thinking loop, effectively distinguishes whether an object is moving.

\begin{table}[t]\footnotesize 
\caption{\textbf{Ablation study} on Davis validation set.}
\vspace{-0.2cm}
\setlength{\tabcolsep}{1.5mm}
\renewcommand\arraystretch{1.0}
\label{tab:Ablation study}
\begin{tabular}{c|l|ccc}
\toprule
Exp.         & Method  & $\mathcal{J\&F}$ ↑  & $\mathcal{J}$ ↑ & $\mathcal{F}$ ↑ \\\hline\hline
\multirow{2}{*}{(a)} & Ours (w/o feature aggregation)       & 90.5            & 87.9            & 93.1          \\
                   & Ours (full, w/ featue aggregation) & \textbf{92.5}   &  \textbf{90.4}  & \textbf{94.6}           \\ \hline
\multirow{2}{*}{(b)} & Ours (w/o deep thinking loop)          & 92.0            & 89.7            & 94.2          \\
                   & Ours (full, w/ deep thinking loop)   & \textbf{92.5}   & \textbf{90.4}   & \textbf{94.6}           \\ \hline
\multirow{3}{*}{(c)} & LISA \cite{lai2024lisa} w/o fine-tuning     & 22.8 & 21.0 & 21.9    \\
                    & LISA \cite{lai2024lisa} w/ fine-tuning    & 70.1 & 69.4 & 69.8    \\
                   & Ours (MovSAM)           & \textbf{92.5}   & \textbf{90.4}   & \textbf{94.6} \\ \bottomrule
\end{tabular}
\vspace{-0.8cm}
\end{table}

\vspace{-0.2cm}
\section{Disscussion}
\vspace{-0.2cm}
MovSAM introduces a novel solution to single-image moving object segmentation by integrating MLLM, SAM, and VLM. While MovSAM utilizes established models, its novelty resides in the effective combination of these components and the introduction of deep thinking for moving object segmentation. MovSAM can serve as a basis for many tasks in robots that usually require multiple frames to solve. For example, with the support of single-image moving object segmentation, single-image optical flow estimation may arise new technical routes. We consider that single-image MOS bridges the gap of the multi-frame MOS task. It is worth to be studied independently, as it helps in the field of recovering temporal information from single-image.

As the results in TABLE \ref{tab:Ablation study} and Fig. \ref{fig:LISA}, language-guided segmentation or open-vocabulary segmentation methods \cite{luo2024emergent, lai2024lisa} still fall short in addressing the challenges of image moving object segmentation. These generalized methods may neglect specific challenges unique to moving object segmentation, such as motion illusions. In contrast, the multiple rounds of deep thinking of MovSAM can fully explore the relationships between objects. In addition, the deep thinking of MovSAM can be migrated to other new tasks that require the ability of logic reasoning from large models, such as cross-modal image-point cloud tasks that consider the interrelationships of objects in the scene.

The inference time of MovSAM is approximately 0.3 seconds. Furthermore, we anticipate that ongoing hardware improvements and optimization techniques for large models will significantly reduce this latency. We believe that MovSAM can stimulate interest across related domains and encourage collaborative efforts to develop improved solutions for this emerging task.

\vspace{-0.2cm}
\section{Conclusion}
\vspace{-0.2cm}
For moving object segmentation, we focus on single-image moving object segmentation. In this paper, we introduce MovSAM, the first framework dedicated to segment moving object in a single image. MovSAM builds a deep thinking pipeline to logically reason about the image for moving object segmentation. Evaluations demonstrate that MovSAM achieves state-of-the-art performance on various MOS datasets with single-image. The deep thinking ability of MovSAM enables it to excel in both autonomous driving and real-world scenes. Furthermore, MovSAM holds promise for enabling robots to better interpret motion intentions and for enhancing system robustness in situations like sensor failures. We anticipate that MovSAM will encourage further research into single-image moving object segmentation and inspire advancements in related areas.

\normalem
\bibliographystyle{IEEEtran}  
\bibliography{root} 

\begin{thebibliography}{10}
\providecommand{\url}[1]{#1}
\csname url@rmstyle\endcsname
\providecommand{\newblock}{\relax}
\providecommand{\bibinfo}[2]{#2}
\providecommand\BIBentrySTDinterwordspacing{\spaceskip=0pt\relax}
\providecommand\BIBentryALTinterwordstretchfactor{4}
\providecommand\BIBentryALTinterwordspacing{\spaceskip=\fontdimen2\font plus
\BIBentryALTinterwordstretchfactor\fontdimen3\font minus \fontdimen4\font\relax}
\providecommand\BIBforeignlanguage[2]{{%
\expandafter\ifx\csname l@#1\endcsname\relax
\typeout{** WARNING: IEEEtran.bst: No hyphenation pattern has been}%
\typeout{** loaded for the language `#1'. Using the pattern for}%
\typeout{** the default language instead.}%
\else
\language=\csname l@#1\endcsname
\fi
#2}}

\bibitem{xie2024moving}
J.~Xie, C.~Yang, W.~Xie, and A.~Zisserman, ``Moving object segmentation: All you need is sam (and flow),'' \emph{arXiv preprint arXiv:2404.12389}, 2024.

\bibitem{cho2023treating}
S.~Cho, M.~Lee, S.~Lee, C.~Park, D.~Kim, and S.~Lee, ``Treating motion as option to reduce motion dependency in unsupervised video object segmentation,'' in \emph{Proceedings of the IEEE/CVF winter conference on applications of computer vision(WACV)}, 2023, pp. 5140--5149.

\bibitem{kirillov2023segment}
A.~Kirillov, E.~Mintun, N.~Ravi, H.~Mao, C.~Rolland, L.~Gustafson, T.~Xiao, S.~Whitehead, A.~C. Berg, W.-Y. Lo, \emph{et~al.}, ``Segment anything,'' in \emph{Proceedings of the IEEE/CVF International Conference on Computer Vision (ICCV)}, 2023, pp. 4015--4026.

\bibitem{guo2025deepseek}
D.~Guo, D.~Yang, H.~Zhang, and J.~Song, ``Deepseek-r1: Incentivizing reasoning capability in llms via reinforcement learning,'' \emph{arXiv preprint arXiv:2501.12948}, 2025.

\bibitem{voigtlaender2019feelvos}
P.~Voigtlaender, Y.~Chai, F.~Schroff, H.~Adam, and B.~Leibe, ``Feelvos: Fast end-to-end embedding learning for video object segmentation,'' in \emph{Proceedings of the IEEE/CVF conference on computer vision and pattern recognition (CVPR)}, 2019, pp. 9481--9490.

\bibitem{koh2017primary}
Y.~J. Koh and C.-S. Kim, ``Primary object segmentation in videos based on region augmentation and reduction,'' in \emph{2017 IEEE conference on computer vision and pattern recognition (CVPR)}.\hskip 1em plus 0.5em minus 0.4em\relax IEEE, 2017, pp. 7417--7425.

\bibitem{wang2024sam}
H.~Wang, P.~K.~A. Vasu, F.~Faghri, R.~Vemulapalli, M.~Farajtabar, S.~Mehta, M.~Rastegari, O.~Tuzel, and H.~Pouransari, ``Sam-clip: Merging vision foundation models towards semantic and spatial understanding,'' in \emph{Proceedings of the IEEE/CVF Conference on Computer Vision and Pattern Recognition (CVPR)}, 2024, pp. 3635--3647.

\bibitem{lai2024lisa}
X.~Lai, Z.~Tian, Y.~Chen, Y.~Li, Y.~Yuan, S.~Liu, and J.~Jia, ``Lisa: Reasoning segmentation via large language model,'' in \emph{Proceedings of the IEEE/CVF Conference on Computer Vision and Pattern Recognition}, 2024, pp. 9579--9589.

\bibitem{wang2022image}
W.~Wang, H.~Bao, L.~Dong, J.~Bjorck, Z.~Peng, Q.~Liu, K.~Aggarwal, O.~K. Mohammed, S.~Singhal, S.~Som, \emph{et~al.}, ``Image as a foreign language: Beit pretraining for all vision and vision-language tasks,'' \emph{arXiv preprint arXiv:2208.10442}, 2022.

\bibitem{dubey2024llama}
A.~Dubey, A.~Jauhri, A.~Pandey, A.~Kadian, A.~Al-Dahle, A.~Letman, A.~Mathur, A.~Schelten, A.~Yang, A.~Fan, \emph{et~al.}, ``The llama 3 herd of models,'' \emph{arXiv preprint arXiv:2407.21783}, 2024.

\bibitem{ravi2024sam}
N.~Ravi, V.~Gabeur, Y.-T. Hu, R.~Hu, C.~Ryali, T.~Ma, H.~Khedr, R.~R{\"a}dle, C.~Rolland, L.~Gustafson, \emph{et~al.}, ``Sam 2: Segment anything in images and videos,'' \emph{arXiv preprint arXiv:2408.00714}, 2024.

\bibitem{perazzi2016benchmark}
F.~Perazzi, J.~Pont-Tuset, B.~McWilliams, L.~Van~Gool, M.~Gross, and A.~Sorkine-Hornung, ``A benchmark dataset and evaluation methodology for video object segmentation,'' in \emph{Proceedings of the IEEE conference on computer vision and pattern recognition (CVPR)}, 2016, pp. 724--732.

\bibitem{ochs2013segmentation}
P.~Ochs, J.~Malik, and T.~Brox, ``Segmentation of moving objects by long term video analysis,'' \emph{IEEE transactions on pattern analysis and machine intelligence (T-PAMI)}, vol.~36, no.~6, pp. 1187--1200, 2013.

\bibitem{li2013video}
F.~Li, T.~Kim, A.~Humayun, D.~Tsai, and J.~M. Rehg, ``Video segmentation by tracking many figure-ground segments,'' in \emph{Proceedings of the IEEE international conference on computer vision (ICCV)}, 2013, pp. 2192--2199.

\bibitem{lee2022iteratively}
Y.~Lee, H.~Seong, and E.~Kim, ``Iteratively selecting an easy reference frame makes unsupervised video object segmentation easier,'' in \emph{Proceedings of the AAAI Conference on Artificial Intelligence (AAAI)}, vol.~36, no.~2, 2022, pp. 1245--1253.

\bibitem{wang2021swiftnet}
H.~Wang, X.~Jiang, H.~Ren, Y.~Hu, and S.~Bai, ``Swiftnet: Real-time video object segmentation,'' in \emph{Proceedings of the IEEE/CVF Conference on Computer Vision and Pattern Recognition (CVPR)}, 2021, pp. 1296--1305.

\bibitem{schmidt2022d2conv3d}
C.~Schmidt, A.~Athar, S.~Mahadevan, and B.~Leibe, ``D2conv3d: Dynamic dilated convolutions for object segmentation in videos,'' in \emph{Proceedings of the IEEE/CVF winter conference on applications of computer vision(WACV)}, 2022, pp. 1200--1209.

\bibitem{zhang2020unsupervised}
L.~Zhang, J.~Zhang, Z.~Lin, R.~M{\v{e}}ch, H.~Lu, and Y.~He, ``Unsupervised video object segmentation with joint hotspot tracking,'' in \emph{Computer Vision--ECCV 2020: 16th European Conference (ECCV), Glasgow, UK, August 23--28, 2020, Proceedings, Part XIV 16}.\hskip 1em plus 0.5em minus 0.4em\relax Springer, 2020, pp. 490--506.

\bibitem{ren2021reciprocal}
S.~Ren, W.~Liu, Y.~Liu, H.~Chen, G.~Han, and S.~He, ``Reciprocal transformations for unsupervised video object segmentation,'' in \emph{Proceedings of the IEEE/CVF conference on computer vision and pattern recognition (CVPR)}, 2021, pp. 15\,455--15\,464.

\bibitem{yang2021learning}
S.~Yang, L.~Zhang, J.~Qi, H.~Lu, S.~Wang, and X.~Zhang, ``Learning motion-appearance co-attention for zero-shot video object segmentation,'' in \emph{Proceedings of the IEEE/CVF international conference on computer vision (ICCV)}, 2021, pp. 1564--1573.

\bibitem{luo2024emergent}
J.~Luo, S.~Khandelwal, L.~Sigal, and B.~Li, ``Emergent open-vocabulary semantic segmentation from off-the-shelf vision-language models,'' in \emph{Proceedings of the IEEE/CVF Conference on Computer Vision and Pattern Recognition (CVPR)}, 2024, pp. 4029--4040.

\end{thebibliography}

\end{document}